\documentclass{article} 
\usepackage{iclr2027_conference,times}

\usepackage{amsmath,amsfonts,bm}

\def\eqref#1{equation~\ref{#1}}

\def\1{\bm{1}}

\DeclareMathAlphabet{\mathsfit}{\encodingdefault}{\sfdefault}{m}{sl}
\SetMathAlphabet{\mathsfit}{bold}{\encodingdefault}{\sfdefault}{bx}{n}

\usepackage{graphicx}
\usepackage{booktabs}
\usepackage{xcolor}
\usepackage{wrapfig}
\newcommand{\red}[1]{\textcolor{black}{#1}}
\usepackage{caption}
\usepackage{multirow}
\usepackage{tikz}
\usepackage[normalem]{ulem}
\usepackage{colortbl}
\usetikzlibrary{positioning,arrows.meta}

\usepackage{hyperref}
\usepackage{url}

\title{World Agent: Can Language Models Keep a World Running?}

\author{%
Weixing Chen\textsuperscript{1,*}, Weipeng Zhang\textsuperscript{5,*}, Nan An\textsuperscript{1}, Yang Liu\textsuperscript{1,3,4,$\dagger$}, Liang Lin\textsuperscript{1,2,3,4} \\
\texttt{\{chenwx228,annan8\}@mail2.sysu.edu.cn}, \texttt{202364871291@mail.scut.edu.cn} \\
\texttt{liuy856@mail.sysu.edu.cn}, \texttt{linliang@ieee.org} \\
\textsuperscript{$*$}Equal contribution. \quad \textsuperscript{$\dagger$}Corresponding author. \\
\textsuperscript{1}Sun Yat-sen University \quad \textsuperscript{2}Peng Cheng Laboratory \\
\textsuperscript{3}Guangdong Key Laboratory of Big Data Analysis and Processing \\
\textsuperscript{4}X-Era AI Lab \quad \textsuperscript{5}South China University of Technology
}

\iclrpreprintcopy 
\begin{document}

\maketitle

\begin{abstract}
World models are moving from generating realistic frames to generating playable worlds, yet whether a delivered world can keep running is not tested anywhere. Existing evaluations stop at generation, at delivery, or at single-step transitions, and each stops at a different point along the way. Correct local state transitions or intermediate outcomes do not guarantee a correctly organized causal event flow. We propose the \textbf{world agent} task, which moves the evaluation point of world generation from the moment of delivery to the continued operation that follows. In this task, a model is not asked to generate a world. It is held responsible for keeping the world running, which requires coordinating events and carrying forward their consequences to constrain subsequent evolution. We instantiate the task in \textbf{WorldAgent-Benchmark} with two complementary tracks. In the maintenance track, the model must ground the events of a continuous narrative into correct transitions of the explicit world state while respecting causal, temporal, and concurrency constraints. In the deduction track, the model must predict how the world will evolve under partial observations and act toward a goal. The maintenance track combines LLM-assisted semantic judgments with programmatic validation and scoring, while the deduction track is evaluated entirely programmatically. Individual judgments are auditable against world states and execution logs, and scores can be recomputed from the saved judgments and execution records. Across 8 models, scores decline steadily as pre-built structure is removed from the world, and causal-relation checking is the weakest component for every model. The benchmark makes the continued operation of a world measurable and distinguishes local completion from failures in event organization. Code and dataset will be released on \url{https://github.com/HCPLab-SYSU/WorldAgent-Benchmark}.
\end{abstract}

\section{Introduction}

Generative world models have advanced from realistic frame synthesis \cite{ha2018worldmodels,sora2024} to playable environment generation \cite{bruce2024genie,valevski2025gamengen,genie2,oasis}, and coding agents have been tasked with building complete games from natural language \cite{zhang2025vgamegym,luo2026gamecraftbench}. The first one is evaluated mainly on visual fidelity and short rollouts, where state consistency can degrade as interactions lengthen \cite{genie2,oasis}. The second is evaluated at the moment of delivery, with no measurement of what follows \cite{zhang2025vgamegym,hu2024gamegen,todd2024gavel}. In both cases the evaluation stops at creation. Whether a delivered world can sustain correct operation over time has remained unmeasured.

\begin{figure}
    \centering
    \includegraphics[width=0.98\linewidth]{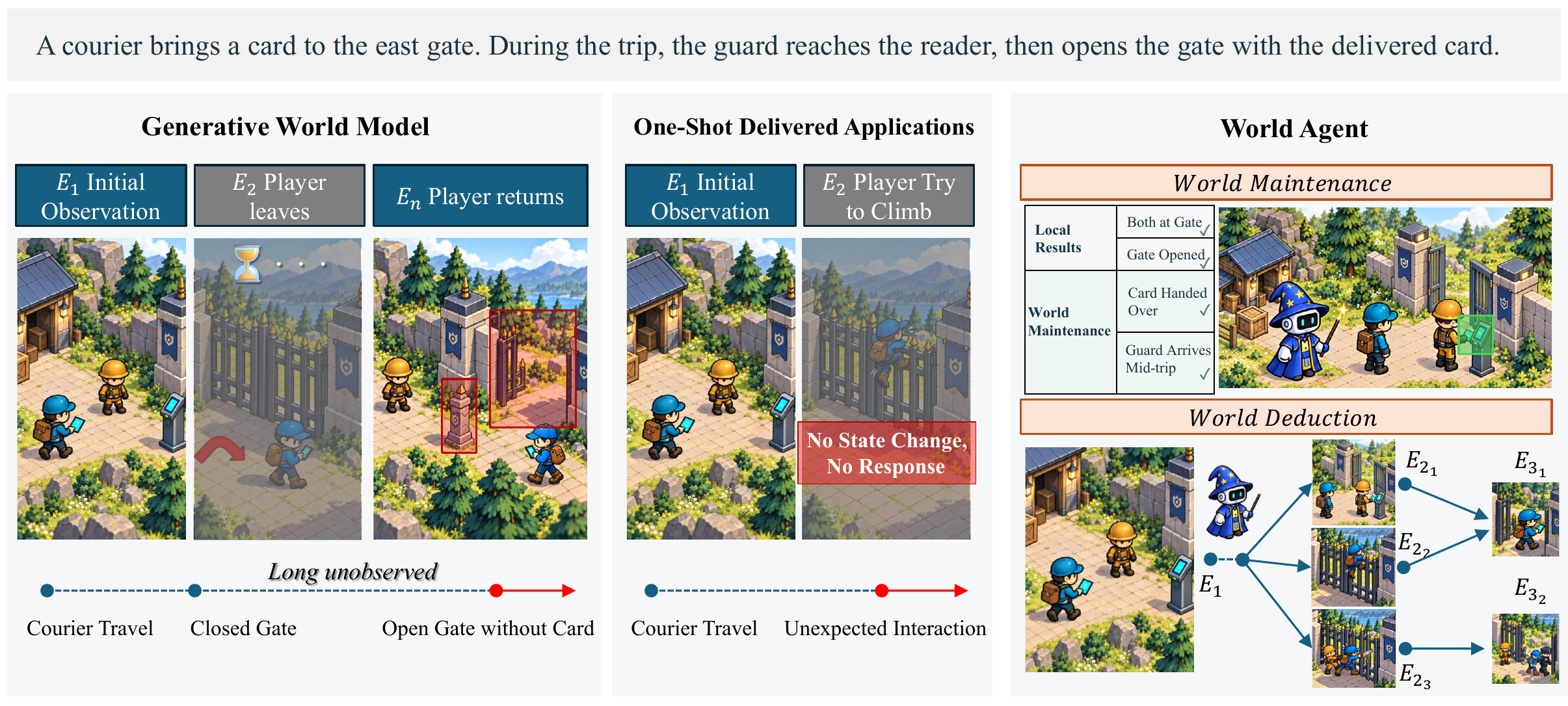}
    \caption{Generative world models and one-shot delivered applications both suffer persistent-state failures that remain unmeasured. The world agent task holds a model responsible for keeping an explicit world running through a shared tool interface, with a maintenance track that grounds events into state updates and a deduction track that predicts evolution from partial observations.}
    \label{fig:overview}
\end{figure}

Sustained operation is fundamentally a problem of state. 
In the scene of Figure~\ref{fig:overview}, a player acting as a courier carries a card to the east gate, and a guard reaches the reader and opens the gate with the delivered card. If the player leaves and returns much later, a generative world model may show the gate open without the card, since nothing has maintained what happened in the player's absence. If the player attempts an unexpected action such as climbing the gate, a one-shot delivered application may produce no state change and no response, having been validated only at delivery. In both cases the world state has failed to persist. These failures are experienced directly by the player, yet the world is not kept running by the player. The courier and the guard act inside the world, while its explicit state is maintained by a separate agent, through which every event is grounded into persistent state updates over a shared tool interface.

Yet persistence alone is not enough. 
Each event must be grounded into the right state transitions while honoring causal dependencies, temporal ordering, duration, and concurrency, and every consequence must be propagated forward so that later evolution remains constrained. Correct local outcomes therefore do not imply a correctly organized event flow. Two characters may both reach the gate and the gate may end up open, yet the implicit handover could be omitted, a dependency reversed, or an event misplaced outside its required interval. This requirement extends well beyond world-model video generation, and any agent expected to act over long horizons in a persistent environment must first be able to maintain its state.

To make sustained operation measurable across diverse worlds requires a shared protocol through which a model can operate different explicit worlds and a shared target for scoring correctness. Such a protocol has been unavailable because explicit-state systems \cite{cote2018textworld,hu2025text2world} have typically been built for a single domain, but large language model agents with generic tool interfaces have now made one practical. The same class of tools can be used to read and write the state of different explicit worlds, and a shared runtime can host cases from multiple domains. With explicit state, execution logs, and a shared interface, every event grounding and every causal, temporal, or concurrency relation can be checked against explicit contracts, so that correct local outcomes can be distinguished from operations that occur without the right prerequisites or in the wrong relation to other operations, rather than being left to a reader's impression of textual coherence.

We propose the world agent task, in which the evaluation point of world generation is moved from the moment of delivery to the continued operation that follows. A model is not asked to generate a world but is held responsible for keeping it running. Just as agent benchmarks made it measurable whether a model can get things done \cite{liu2024agentbench,zhou2024webarena,xie2024osworld,jimenez2024swebench}, the world agent task makes it measurable whether a world can be kept running. The task is posed in two complementary forms. In maintenance tasks, a continuous narrative stream is received, and each round's events must be reactively grounded into correct transitions of the explicit world state. In deduction tasks, an initial state and an incomplete description of the dynamics are given, and the model must predictively reason about how the world will evolve. Figure~\ref{fig:overview} illustrates both forms.

WorldAgent-Benchmark instantiates this task with two tracks. The maintenance track contains 31 cases across three difficulty tiers, in which pre-built structure is progressively removed. All world structure is provided in the easiest tier, while only a map is given in the hardest tier and the types, items, and rules must be constructed by the model. Causal, temporal, and concurrency relations are checked by programs, the remaining semantic judgments are made with LLM assistance and examined item by item in an isolated sandbox, and all scoring is programmatic. The deduction track contains 6 cases under two information conditions, in which predictions are compared field by field against the exact evolution computed independently by the engine, with no language-model judge. Every judgment is logged, and all checks and score aggregation can be recomputed end to end.

Eight models are evaluated, and a reproducible difficulty gradient is found, with scores declining steadily as pre-built structure is removed. Causal-relation satisfaction is the weakest component for every model and degrades monotonically with difficulty. This pattern distinguishes local completion from failures in event organization, so sustained operation is assessed not merely by whether a desired state appears but by how the world reaches it. Our contributions are summarized as follows.

\begin{itemize}
  \item 
  We propose the world agent task, which shifts the evaluation point of world generation from delivery to continued operation, posed in two complementary forms, maintenance and deduction.

  \item We build an evaluation protocol in which every judgment is programmatic or item-by-item auditable, and instantiate it in WorldAgent-Benchmark with 37 cases across two tracks and three construction tiers.

  \item Whether models can keep a world running remains unmeasured. Across 8 models we find a reproducible difficulty gradient and a dominant failure mode in causal-relation errors.
\end{itemize}
\section{Related Work}
\label{sec:related}

\paragraph{World Models}
\label{sec:related-worldmodels}
Generative world models synthesize playable environments frame by frame \cite{ha2018worldmodels,sora2024,bruce2024genie,valevski2025gamengen,genie2,oasis}, and recent work has begun to probe their long-horizon stability, though the checks remain visual and geometric \cite{xu2026worldroambench}. A second line represents world state explicitly instead of rendering it. Tracking state changes through fixed narratives that the model only reads is an older problem in language understanding \cite{weston2015babi,dalvi2018propara}. Recent explicit-state work records spatial and appearance state \cite{xiao2025worldmem,momennejad2023cogeval,yang2024vsibench}, tests transitions as single-step prediction \cite{wang2024textworld}, or generates symbolic world models in one shot \cite{hu2025text2world}. The GEST-Engine \cite{cudlenco2026gest}, WiA-LLM \cite{sui2025wiallm}, and the Programmable World Model \cite{huang2026pwm} maintain explicit state through a language model that proposes changes for a program backend to apply, but they target generation or training rather than benchmarking. The closest evaluation is WSE-bench \cite{chen2026wsebench}, which checks whether a story stays consistent as its world evolves, yet its canon is reconstructed from the model's own narrative and every judgment comes from a language-model judge. Across this line the model either generates the world or reads along with it, and evaluation stops at single steps, short horizons, or the model's own account of what happened. We instead hand the model an already delivered world with explicit state and score what it becomes over long horizons against ground truth frozen before any run.

\paragraph{Coding Agents and Game Generation}
\label{sec:related-coding}
Agent benchmarks \cite{liu2024agentbench,zhou2024webarena,xie2024osworld,jimenez2024swebench,yao2024taubench} and persistent game or embodied environments \cite{cote2018textworld,hausknecht2020jericho,urbanek2019light,shridhar2021alfworld,hafner2022crafter,wang2023voyager,liu2025aligning} score what an agent achieves while a fixed engine maintains the world, so the correctness of the world itself is never evaluated. Game generation has been adopted as a stress test for coding agents \cite{zhang2025vgamegym,hu2024gamegen,todd2024gavel,luo2026gamecraftbench}, and these benchmarks score a game at the moment it is delivered. The closest work in this line is GameLogicBench \cite{che2026gamelogicbench}, where coding agents implement gameplay mechanics in Godot projects and a deterministic judge checks tick-level state assertions. It moves the evaluation point from delivery into execution, but the object of evaluation is still the code the agent writes, and the world itself is maintained by the engine and assumed to run correctly. A recent survey of AI for games \cite{luo2026aiforgames} confirms that none of its six roles is responsible for a world's continued operation. We swap this assignment of responsibility. The model writes no game code and plays no character, but is responsible for keeping a delivered world running, and the benchmark scores exactly that.
\section{The World Agent Task}
\label{sec:task}

We formalize maintaining causal event flows through a task instance $\mathcal T$ and a timed execution trace $\tau$ generated by its execution mechanism, shown as\ref{eq:event-flow}.
Here $s_0$ is the initial world state, $F$ is the execution mechanism, and $X$ is the instance-specific inputs. The finite specification $\mathcal C$ constrains event realization, rule-consistent changes, causal dependencies, ordering, duration, and concurrency. The trace has $K$ records: $\ell_k$ records the k-th operations and $s_k$ is the resulting state. Each $c(\tau)\in\{0,1\}$ indicates whether one requirement holds; $\models$ denotes satisfaction of all requirements. These may span the trace, so earlier consequences constrain later evolution.

\vspace{-10pt}
\begin{equation}
\begin{aligned}
\mathcal T &= (s_0,F,X,\mathcal C),\qquad
\tau = (s_0,\ell_1,s_1,\ldots,\ell_K,s_K),\\
\tau\models\mathcal C
&\ \Longleftrightarrow\
c(\tau)=1\quad\text{for every }c\in\mathcal C .
\end{aligned}
\label{eq:event-flow}
\end{equation}

We evaluate two complementary aspects of this capability. \textbf{World maintenance} realizes and organizes the required event flow; \textbf{world deduction} predicts autonomous and action-conditioned evolution. They instantiate $F$ through a maintenance interface $F_A$ or a deduction protocol with fixed dynamics $f$, respectively, and are scored separately.

\subsection{World Maintenance}
\label{sec:task-maintenance}

In maintenance, $X=x_{1:T}$ is a narrative delivered over $T$ rounds, and $\mathcal C$ is organized into hidden round contracts $C_{1:T}$. Each $C_t$ specifies explicit and necessary implicit events, required structures, and within- or cross-round constraints. Before interaction $j$ of round $t$, the model has received only $x_{\le t}$. Let $\tau_{t,j}$ be the current execution prefix and $h_{t,j}$ the prior tool-request/response history, carried across rounds. The model's policy $\pi$ selects a tool request $a_{t,j}$:
\begin{equation}
\begin{aligned}
a_{t,j}&\sim\pi(x_{\le t},h_{t,j}),\qquad
(\tau_{t,j+1},o_{t,j})=F_A(\tau_{t,j},a_{t,j}),\\
h_{t,j+1}&=h_{t,j}\oplus(a_{t,j},o_{t,j}).
\end{aligned}
\label{eq:maintenance}
\end{equation}
Here $o_{t,j}$ is the query result or operation feedback, and $\oplus$ appends the request--response pair. Queries update the interaction history without extending the world trace. Only the engine writes world state, checking structural validity and atomic writes without supplying missing events or enforcing semantic prerequisites. The model must prepare objects and capabilities and explicitly realize and schedule behaviors across the world, rather than act only as one character.

Let $b_t$ be the last execution-record index at round $t$'s frozen boundary, and $\tau_{\le b_t}$ the trace prefix ending there. Complete satisfaction of the maintenance task means
\begin{equation}
\mathrm{Success}_A
\ \Longleftrightarrow\
\bigwedge_{t=1}^{T}\bigl[\tau_{\le b_t}\models C_t\bigr].
\label{eq:maintenance-success}
\end{equation}
Contracts assess entire prefixes, which later operations cannot repair. This ideal condition is scored item by item in Section~\ref{sec:maintenance}, not as an additional all-or-nothing metric. Round initialization and interaction limits appear in Appendix~\ref{app:runtime}.

\subsection{World Deduction}
\label{sec:task-deduction}

Deduction tests whether a model can predict world evolution under fixed deterministic dynamics $f$, including autonomous processes and other actors. We cross \textbf{Open/Naming} information conditions with \textbf{M1/M2} rollout modes, forming a $2\times2$ design. Open provides more explicit process or action rules, whereas Naming leaves details unspecified. M1 predicts autonomous evolution without state feedback; M2 couples prediction with actions and feedback.

At prediction point $j$, let $h_j$ be the model's visible history of disclosed inputs, prior predictions, and permitted feedback. A query $q_j$ specifies what evolution to predict, under which actions and over what interval. The model's policy $\pi$ produces a prediction $\hat y_j$; let $y_j$ denote the answer implied by the actual dynamics $f$. A correct prediction satisfies
\begin{equation}
\hat y_j\sim\pi(h_j,q_j),\qquad \hat y_j=y_j.
\label{eq:deduction}
\end{equation}

In \textbf{M1}, the model predicts successive segments of autonomous evolution from the initial public state and disclosed rules, without receiving true intermediate states or correctness feedback. In \textbf{M2}, it controls one actor toward a goal and submits an action together with a prediction of all world changes until the next decision point, before receiving the resulting feedback. It also predicts evolution under a proposed plan and the consequences of alternative past actions.

Predictions describe world evolution under $f$; they do not themselves update the world.

\section{WorldAgent-Benchmark}
\label{sec:benchmark}

\begin{figure}[t]
  \centering
  \includegraphics[width=0.98\linewidth]{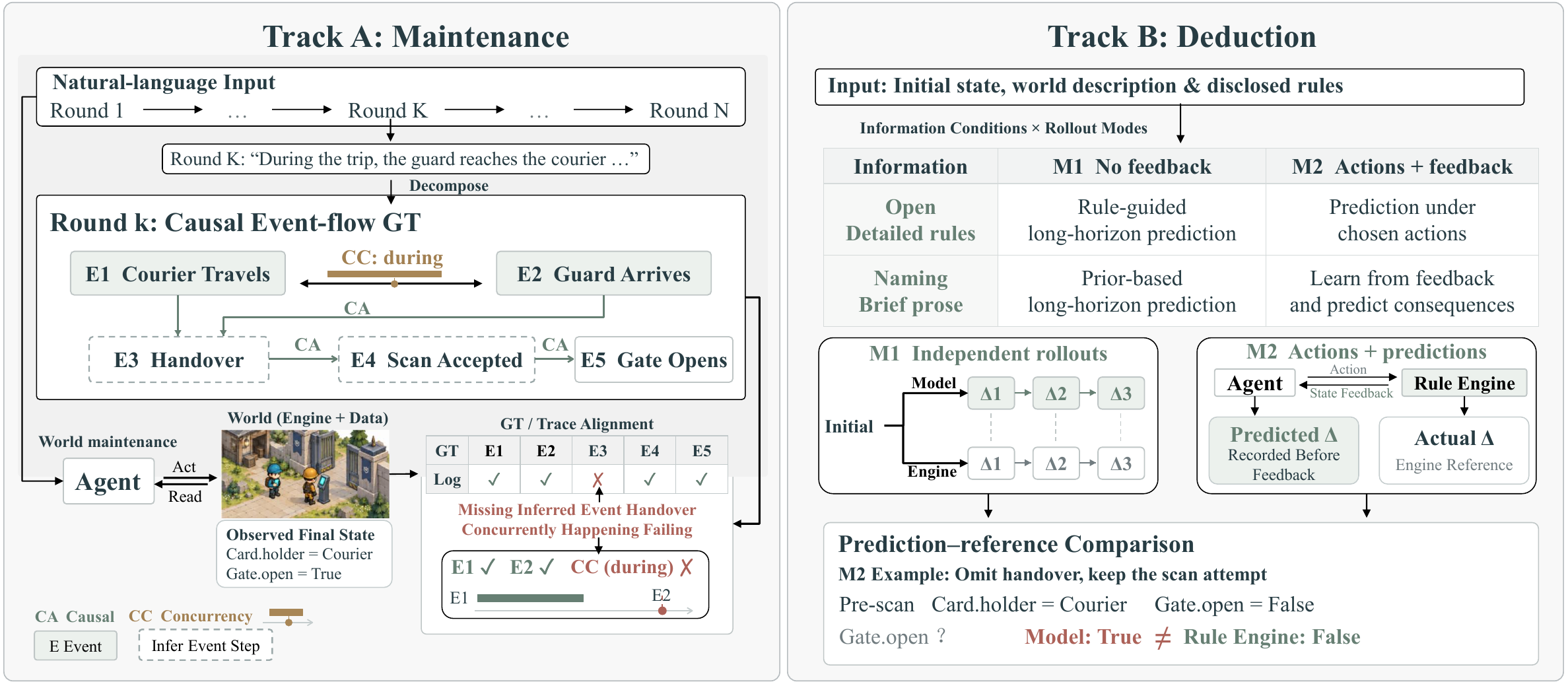}
  \caption{The WorldAgent-Benchmark pipeline. Ground truth for both tracks is authored and frozen before any run, each track runs on the same runtime under one of two task conditions, and scoring reads only frozen artifacts. The maintenance track (left) is scored against per-round contracts by a language-model judge and deterministic programs. The deduction track (right) is scored field by field against the engine's exact evolution, with no language-model judge. Badges at the bottom state the three design commitments.}
  \label{fig:pipeline}
\end{figure}

WorldAgent-Benchmark instantiates maintenance and deduction as separate tracks (Figure~\ref{fig:pipeline}). They share explicit world records and a separation between tested outputs and evaluation criteria, but require different authoring procedures and sources of ground truth.

\subsection{Case Construction and Experimental Settings}
\label{sec:runtime}
\label{sec:authoring}

Each case fixes its specification and evaluation protocol before testing. An episode selects a maintenance difficulty tier, or a deduction information condition and rollout mode; a run executes that episode once with a model. In both tracks, the engine alone writes authoritative world state, but its responsibilities differ as defined in Section~\ref{sec:task}.

For maintenance, authors jointly construct a narrative and its event-flow contracts, including implicit intermediate events and cross-round constraints. They then materialize the narrative inputs and difficulty-specific world data. The model realizes the narrative through world operations; frozen snapshots and execution logs are subsequently assessed against the held-out contracts.

For deduction, authors specify initial conditions, action and process rules, and a zero-intervention reference timeline. Case modules implement these dynamics over a generic deterministic kernel, and validation checks their agreement with the authored specification. Rather than manually labeling every possible intervention, the engine computes the reference rollout for each prediction query. Authoring, validation, and baseline protocols are detailed in Appendix~\ref{app:construction}; runtime interfaces are in Appendix~\ref{app:runtime}.

\subsection{The Maintenance Track}
\label{sec:maintenance}

Each round $t$ has an authored contract $C_t=(E_t,A_t,P_t,R_t)$. $E_t$ is the set of required event nodes: each $e\in E_t$ specifies a behavior and its completion condition, either explicit in the narrative or inferred as a necessary intermediate step. $A_t$ contains availability and use requirements for types, entities, and capabilities; $P_t$ contains probes for conditional prohibitions over declared scopes. $R_t$ contains the causal (CA), temporal (TM), concurrency (CC), and within-chain order requirements checked in that round.

The narrative, map, and contracts remain fixed across three difficulty tiers. \textbf{Easy} supplies types, entities, and capabilities. \textbf{Hard} requires the model to construct capabilities, while types and entities remain provided. \textbf{Extreme} leaves only the map pre-built, requiring the model to construct all three kinds of structure.

An LLM judge uses frozen world states and execution logs to bind E/A items and assess probe violations. An E node must bind one or more successful triggers from its own round whose recorded behavior and effects satisfy its completion condition; a matching final state alone is insufficient. Programs validate the bindings, derive event intervals, and settle $R_t$. Thus, event realization and relation satisfaction are distinct: two events can occur while their relation fails.

For a run at difficulty $d$, every node in $E_t$ enters completion. Let $A^{\mathrm{sc}}_{t,d}$ be the capability/entity requirements assigned to the model and scored in round $t$, and $P^{\mathrm{sc}}_t$ the probes settled there; $\mathrm{sc}$ denotes scored items. Pre-built structures and type-level requirements are audited but excluded from completion. Cross-round A/P items are merged and counted once. For any requirement set $S$, $|S|$ is its item count and $S^+$ its passed subset in this run: in particular, $E_t^+$ contains established events and $(P^{\mathrm{sc}}_t)^+$ contains probes with no violation found in their scopes. We operationalize Eq.~\eqref{eq:maintenance-success} with two item-level scores:
\begin{equation}
\begin{aligned}
\mathrm{completion}_{t,d}
=\frac{|E_t^+|+|(A^{\mathrm{sc}}_{t,d})^+|+|(P^{\mathrm{sc}}_t)^+|}
{|E_t|+|A^{\mathrm{sc}}_{t,d}|+|P^{\mathrm{sc}}_t|},
\qquad
\mathrm{relation}_{t,d}=\frac{|R_t^+|}{|R_t|}.
\end{aligned}
\label{eq:maintenance-scores}
\end{equation}
Relation failures do not reduce the event-completion count. We retain component counts and report the two scores separately; tier comparisons must account for their different construction requirements. Detailed binding, probe-scope, interval, cross-round settlement, and empty-denominator rules appear in Appendix~\ref{app:scoring}.

\subsection{The Deduction Track}
\label{sec:deduction}

The $2\times2$ design in Figure~\ref{fig:pipeline} distinguishes four settings. \textbf{Open--M1} tests autonomous rollout from explicit process rules; \textbf{Naming--M1} tests prediction from qualitative descriptions without corrective feedback, making it sensitive to domain priors. \textbf{Open--M2} tests action-conditioned prediction with a disclosed action table; \textbf{Naming--M2} instead provides brief action descriptions and allows missing details to be inferred from interaction feedback. Paired Open/Naming conditions retain the same case dynamics and initial public state. Exact disclosure and feedback rules are given in Appendix~\ref{app:deduction-runtime}.

For each submitted query, the engine independently executes the fixed dynamics under the specified action conditions to obtain $y_j$ in Eq.~\eqref{eq:deduction}. Programs compare the prediction with this reference, without an LLM judge. The primary measure is field-level F1: a match requires both the changed field's path and its resulting value. M1 reports a per-segment prediction curve and an episode-level F1; M2 reports separate F1 scores for action-interval, lookahead, and counterfactual predictions. Interval targets include autonomous changes, not only effects directly caused by the chosen action. Task outcomes and background maintenance are reported separately from prediction accuracy, rather than merged into a total score. Reference construction, aggregation, empty-segment handling, secondary readouts, and baseline definitions are in Appendix~\ref{app:deduction-scoring}.

\paragraph{Auditability.}
The maintenance track uses an LLM judge, which may incorrectly match events or structures or miss violations. Saved judgments and references to world states and execution logs allow individual decisions to be reviewed. With these judgments fixed, scores can be recomputed without calling the LLM again. This reproducibility does not guarantee that the judgments are correct, and rerunning the judge may produce different results.

\section{Experiments}
\label{sec:exp}

\paragraph{Setup.} 
We evaluated 8 models on both tracks of WorldAgent-Benchmark. DeepSeek v4.1 Flash, Qwen-3.8 Flash, and GLM 5.3 Flash ran the full benchmark, covering 93 runs on the maintenance track (31 cases $\times$ 3 difficulty levels) and 120 episodes on the deduction track (6 cases $\times$ 2 modes $\times$ 2 information conditions $\times$ 5 folds). The other five models (Kimi K3, GLM 5.3, Gemini 3.8 Flash, Qwen-3.8-Max-0902 and GPT-5.6 luna) were evaluated on a more discriminative subset of 9 cases selected from the results of the first three models. All runs followed the tool protocol and interaction budget described in Section~\ref{sec:benchmark} without human intervention. Model pairs were compared with a two-sided Wilcoxon signed-rank test at a significance level of $p < 0.05$. Notably, all models are evaluated by default with PhyAgentOS~\cite{liu2026phyagentos}. 

\paragraph{Metrics} The maintenance track reports per-round completion and relation (Section~\ref{sec:maintenance}); we also scan per-round curves for derailment, marked when completion reaches $0.5$ in round one and later falls below $0.2$. The deduction track reports four field-level F1 readouts against the engine-computed evolution (Section~\ref{sec:deduction}): segment F1 in the pure-deduction mode (M1), interval F1 in the closed-loop mode (M2), and lookahead F1, each averaged over the open and naming conditions, and counterfactual F1, merged across information conditions.

\subsection{Main Results}
\label{sec:exp-main}

\begin{table}[!t]
  \caption{Main results on both tracks. The left panel shows the mean per-run score of the nine-case subset evaluated by all 8 maintenance models, with 27 runs per model. The right panel shows field-level F1 averaged over the open and naming conditions, with counterfactual modes merged. Bold entries mark the best result in each column within the upper block. $^{\dagger}$ denote the model working on deepseek harness, while the other is working on PhyAgentOS.}
  \label{tab:main-results}
  \centering
  \scriptsize
  \setlength{\tabcolsep}{12pt}
  \begin{tabular}{@{}l cc cccc@{}}
    \toprule
    & \multicolumn{2}{c}{Maintenance} & \multicolumn{4}{c}{Deduction} \\
    \cmidrule(lr){2-3} \cmidrule(lr){4-7}
    Model & Compl. & Rel. & M1 & M2 & Lookahead & Counterfactual \\
    \midrule
    \rowcolor{blue!10}
    GLM 5.3 Flash & 0.752 & 0.574 & 0.392 & 0.494 & 0.143 & 0.323 \\
    \rowcolor{blue!10}
    Qwen-3.8 Flash & 0.735 & 0.627 & 0.315 & 0.414 & 0.145 & 0.423 \\
    \rowcolor{blue!10}
    DeepSeek v4.1 Flash & 0.744 & 0.580 & 0.395 & 0.522 & 0.159 & 0.466 \\
    DeepSeek v4.1 Flash$^{\dagger}$ & 0.747 & 0.584 & \textbf{0.431} & 0.550 & 0.151 & 0.458 \\
    GPT 5.6 Luna & 0.734 & 0.543 & 0.182 & 0.352 & 0.156 & 0.405 \\
    Gemini 3.8 Flash & 0.704 & 0.512 & 0.238 & 0.547 & \textbf{0.225} & 0.397 \\
    Kimi K3 & 0.709 & 0.546 & 0.377 & 0.543 & 0.148 & \textbf{0.593} \\
    GLM 5.3 & 0.692 & 0.516 & 0.174 & 0.276 & 0.087 & 0.232 \\
    Qwen-3.8-Max-0902 & \textbf{0.813} & \textbf{0.738} & 0.410 & \textbf{0.589} & 0.169 & 0.474 \\
    \bottomrule
  \end{tabular}
\end{table}

Table~\ref{tab:main-results} reports the main results on both tracks. On the nine-case common subset of the maintenance track, completion scores span $0.692$ to $0.813$ across the nine models. Qwen 3.8 Max is the clear leader on both completion ($0.813$) and relation ($0.738$), significantly ahead of every other model (all $p < 0.002$), with Qwen-3.8 Flash second on relation ($0.627$). The remaining models cluster tightly on completion between $0.692$ and $0.752$, led by GLM 5.3 Flash. The DeepSeek v4.1 Flash harness swap (pyos $\rightarrow$ dsh) leaves both scores unchanged ($0.744 \rightarrow 0.747$ and $0.580 \rightarrow 0.584$, both $p \geq 0.37$). \red{This insensitivity indicates that the measured gap reflects model capability rather than harness engineering, and that event organization cannot be recovered by wrapping the same model in a different agent framework.} Completion saturates across most models, whereas relation scores separate them more clearly, since getting the ordering and causality right is the real differentiator.

On the deduction track, M1 scores fall far below M2, showing that models rely on the rule text when stage predictions must be committed in advance and recover only once the loop is closed and each action is followed by the true state change returned by the engine. \red{On M1, the best reading is the dsh re-run of DeepSeek v4.1 Flash ($0.431$).} On M2, Qwen 3.8 Max attains the highest interval average ($0.589$), significantly above the three full-benchmark models (vs.\ Qwen-3.8 Flash $0.414$ and GLM 5.3 Flash $0.494$, both $p < 5\times 10^{-4}$, and vs.\ DeepSeek v4.1 Flash $0.522$, $p = 0.012$). Its relation and closed-loop advantage is consistent with the maintenance-track lead, whereas the smaller Qwen-3.8 Flash remains significantly below DeepSeek and GLM 5.3 Flash on M2. \red{The dsh re-run of DeepSeek v4.1 Flash shows no significant change on the deduction track either ($0.395 \rightarrow 0.431$ on M1 and $0.522 \rightarrow 0.550$ on M2).} Lookahead is a shared weakness, with all models below $0.25$, \red{Gemini 3.8 Flash alone above $0.2$ ($0.225$)}, and none able to predict the world state 48 time units ahead reliably. Counterfactual reasoning is dissociated from pure deduction. GLM 5.3 Flash matches DeepSeek on M1, yet its counterfactual score of $0.323$ is the lowest among the \red{four} models with multi-fold or harness-verified data, while Kimi K3 attains the highest counterfactual reading ($0.593$) alongside high M1 and M2 scores, though from a single fold that we treat as preliminary. \red{Taken together, these capability profiles are unevenly distributed. The strongest completion does not translate into the strongest relation satisfaction, and the best lookahead and counterfactual readings are attained by different models than the best overall performers. Sustained operation therefore does not appear to be a single skill that scales with general model quality, and the components probed by the benchmark appear to be acquired and lost independently.}

\begin{wrapfigure}{r}{0.42\linewidth}
  \centering
  \vspace{-70pt}
  \includegraphics[width=\linewidth]{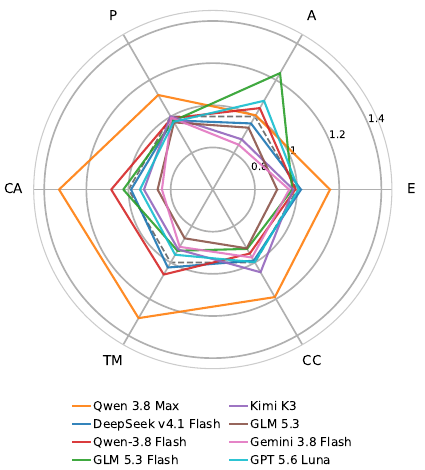}
  \caption{Component strengths on the maintenance track
  . Per-component pass rates are averaged over the three difficulty tiers and normalized by the cross-model mean, with the dashed ring marking $1.0$. Per-model radars are given in Appendix~\ref{app:component-radar}.}
  \label{fig:component-radar}
  \vspace{-10pt}
\end{wrapfigure}

\subsection{Maintenance Track}
\label{sec:exp-maintenance}

\begin{figure}[!t]
  \centering
  \includegraphics[width=\linewidth]{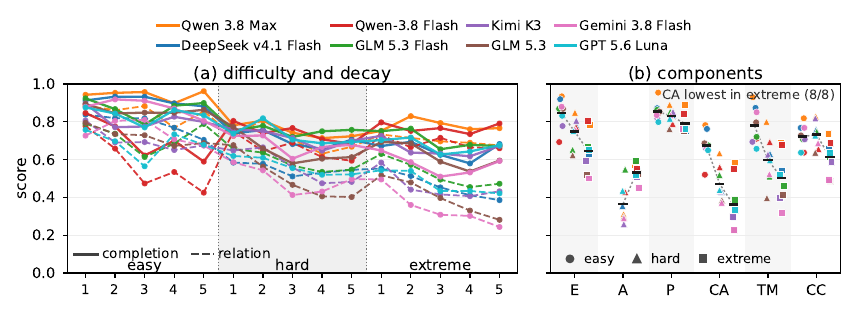}
  \caption{Maintenance track. Panel (a) shows per-round completion (solid lines) and relation scores (dashed lines) broken down by difficulty tier. Panel (b) reports component pass rates, where in each component the three markers denote the easy (circle), hard (triangle), and extreme (square) tiers, and short black ticks mark the cross-model mean within each tier.}
  \label{fig:maintenance-main}
\end{figure}

Figure~\ref{fig:maintenance-main}(a) places the difficulty gradient and the round-by-round decay in the same plot, with the x axis showing the round within a tier, solid lines completion, and dashed lines relation scores. On the nine-case common grid, all \red{eight} models degrade from the easy to the extreme tier, and the relative completion drop from the first round of easy to the last round of extreme reaches \red{$34.8\%$} for DeepSeek v4.1 Flash ($0.907 \rightarrow 0.591$), \red{$32.7\%$} for Gemini 3.8 Flash ($0.879 \rightarrow 0.592$), \red{$22.8\%$} for GPT 5.6 Luna ($0.876 \rightarrow 0.676$), and \red{$18.7\%$} even for Qwen 3.8 Max ($0.942 \rightarrow 0.766$). The only model that resists the gradient is \red{Qwen-3.8 Flash}, whose completion falls by \red{only $10.9\%$} ($0.823 \rightarrow 0.733$), an ordering that survives the relative measure and is therefore not a floor effect of its lower starting point. Its relation score falls only from \red{$0.721$ to $0.625$}, flattening the difficulty ordering of the other \red{seven}. Within every tier both scores decline over rounds and the curves are nearly parallel across models, so the decay is shared rather than an artifact of any single model.

Figure~\ref{fig:maintenance-main}(b) breaks the scores down into six components, from which two patterns emerge. First, in the easy tier the weakest component is CA (causal checking) for \red{seven of the eight} models, the exception being GPT 5.6 Luna, whose weakest component is CC. In the extreme tier CA is the weakest for all \red{eight} models, and its cross-model mean drops from \red{$0.683$} to \red{$0.356$}, \red{the steepest decline among components and the lowest extreme-tier mean}. Second, in the hard tier the weakest component shifts from CA to A (construction) for \red{seven} models, the exception being GLM 5.3 Flash, whose weakest remains CA. Once rules must be authored from scratch, construction failures become more fatal than causal failures. Consistent with this, GLM 5.3 Flash attains the strongest hard-tier construction score ($0.506$), followed by GPT 5.6 Luna ($0.435$), while the remaining \red{six} models stay below $0.35$. \red{A possible explanation is the code-heavy pretraining shared by these models, which we leave to future work to verify.}

\red{Figure~\ref{fig:component-radar} summarizes component strengths as each model's pass rate relative to the cross-model mean. Qwen 3.8 Max is the only model above average on every component, and by a wide margin on CA ($1.33\times$), TM ($1.30\times$), and CC ($1.19\times$). The other models each have at most one distinctive strength, construction (A) for GLM 5.3 Flash ($1.24\times$) and GPT 5.6 Luna ($1.09\times$), causal and temporal checking (CA, TM) for Qwen-3.8 Flash, and concurrency (CC) for Kimi K3. DeepSeek v4.1 Flash, Gemini 3.8 Flash, and GLM 5.3 stay within $\pm 8\%$ of the mean on every component, matching their flat profiles in Figure~\ref{fig:maintenance-main}(b).}

\subsection{Deduction Track}
\label{sec:exp-deduction}

\begin{figure}[t]
  \centering
  \begin{minipage}[t]{0.50\linewidth}
    \centering
    \scriptsize\setlength{\tabcolsep}{8pt}\renewcommand{\arraystretch}{1.4}%
    \raisebox{-\height}{%
    \begin{tabular}{@{}lccc@{}}
      \toprule
      Model & M1 field & M1 fired & M2 int. \\
      \midrule
      Qwen-3.8-Max-0902        & 0.67/0.15 & 0.63/0.50 & 0.64/0.53 \\
      DeepSeek v4.1 Flash & 0.68/0.11 & 0.62/0.41 & 0.56/0.48 \\
      Qwen-3.8-Flash      & 0.56/0.07 & 0.54/0.36 & 0.50/0.33 \\
      GLM 5.3 Flash       & 0.71/0.07 & 0.63/0.37 & 0.56/0.43 \\
      Kimi K3 & 0.72/0.04 & 0.53/0.38 & 0.63/0.45 \\
      GLM 5.3             & 0.24/0.10 & 0.53/0.42 & 0.40/0.16 \\
      GPT 5.6 Luna        & 0.26/0.07 & 0.57/0.33 & 0.42/0.32 \\
      Gemini 3.8 Flash    & 0.44/0.03 & 0.71/0.28 & 0.57/0.53 \\
      \bottomrule
    \end{tabular}}%
    \captionof{table}{Field-level F1 by model, with each cell reporting the open and naming conditions as \textbf{open/naming}. M1 field, M1 fired, and M2 int. denote the pure-deduction segment F1, the rule-firing F1, and the closed-loop interval F1, respectively.}
    \label{tab:deduction-main}
  \end{minipage}\hfill
  \begin{minipage}[t]{0.47\linewidth}
    \centering
    \raisebox{-\height}{\includegraphics[width=\linewidth]{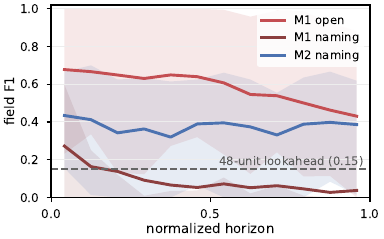}}
    \vspace{-10pt}
    \captionof{figure}{Per-segment F1 vs.\ normalized horizon, pooled over the seven models, for M1 open, M1 naming, and M2 naming (shaded interquartile bands); dashed line: mean 48-unit lookahead. Pure deduction decays with the horizon; per-step feedback keeps it nearly flat.}
    \label{fig:deduction-horizon}
  \end{minipage}
    \vspace{-15pt}
\end{figure}

Table~\ref{tab:deduction-main} reports field-level F1 on the deduction track under different information conditions. Under pure deduction (M1), switching from the open to the naming condition drops field F1 from $0.56$--$0.72$ among the top models to \red{$0.03$--$0.15$} across all \red{eight} models, showing that models rely heavily on machine-readable rule text rather than genuinely mastering the world dynamics. Rule-firing F1 under the naming condition remains at \red{$0.28$--$0.50$}, roughly \red{three} to ten times the field F1, so models can tell which rules will fire but not the concrete values those rules produce, and the collapse occurs mainly at the value level. Closed-loop feedback in M2 pulls interval F1 back to $0.16$--$0.53$, yet an open--naming gap of $7\%$--$34\%$ remains among the six multi-fold or harness-verified models (Kimi K3: $29\%$). \red{The recovered fraction varies by model, from about $10\%$ of the naming-condition gap for Gemini 3.8 Flash to about $34\%$ for Qwen-3.8 Flash, with DeepSeek v4.1 Flash and GLM 5.3 Flash in between at $15\%$ and $21\%$.} The three full-benchmark models separate here, with Qwen-3.8 Flash at $0.414$ significantly below DeepSeek v4.1 Flash and GLM 5.3 Flash. \red{Qwen 3.8 Max shows the mildest naming collapse (M1 field $0.15$ vs.\ $0.03$--$0.11$) together with the highest M2 intervals ($0.64/0.53$), although its single-fold readout remains preliminary.} Counterfactual scores are not aligned with interval competence. GLM 5.3 Flash reaches $0.49$ on interval F1 but only $0.32$ on counterfactual F1, whereas Qwen-3.8 Flash is nearly flat across the two ($0.42$ vs.\ $0.41$). Predicting how the world evolves and answering what would change under a different action are therefore separable abilities. \red{One caveat concerns the pure-deduction mode. Case-difficulty correlations across models are strong and significant in closed-loop mode ($r = 0.84$--$0.99$) but weak and mostly insignificant under M1, so M1 readouts mix deductive ability with genre-specific priors while closed-loop interval F1 is a cleaner cross-model measure.}

Figure~\ref{fig:deduction-horizon} unfolds the same conclusion along the horizon. Without rule text, M1 naming decays monotonically from $0.27$ to $0.04$, and even under the open condition M1 falls from $0.68$ to $0.44$, indicating that pure deduction error grows with the prediction distance. With closed-loop feedback, M2 naming stays nearly flat, moving only from $0.43$ to $0.39$, and the long-horizon decay is largely suppressed. The 48-unit lookahead, however, hovers around $0.15$, below every other curve in the figure. Models can thus use step-by-step feedback to keep the current state plausible, but they still lack a reliable explicit forward prediction of the world state 48 time units ahead.

\section{Conclusion}

We propose the world agent task, which evaluates world generation by sustained operation rather than the moment of delivery, and assess it with WorldAgent-Benchmark on maintenance and deduction tracks. Scores fall reproducibly as pre-built structure is removed, causal checking is the weakest component, and no model predicts the world state 48 time units ahead with any reliability. Generating a world is therefore not the same as sustaining one, and the task offers a reproducible target for measuring sustained operation and pinpointing the gaps current models must still close.

\bibliography{main}
\bibliographystyle{iclr2027_conference}

\appendix

\section{Case Inventory and Construction Protocol}
\label{app:construction}

\paragraph{Case inventory} The maintenance narratives span genres from disaster rescue and criminal investigation to settlement management and manufacturing, and their scale ranges from a three-round town crisis to a stress case (\texttt{Ark\_001}) with \red{287} scored events over a map of \red{395} tiles. The deduction settings range from a town on the eve of an outbreak to a leaking space station and an espionage network.

\paragraph{Maintenance cases}

Authors jointly develop a complete narrative and its decomposition into event nodes, structure requirements, conditional prohibitions, and event relations. The ground-truth document set contains a story overview with all inputs, a map decomposition, and per-round contracts in fixed tables. Once reviewed, this specification is frozen for the case version; runtime inputs, per-tier initial data, round presets, and presentation assets must agree with it. LLMs may assist drafting, but semantic review remains separate from mechanical validation. Every scoring requirement must trace to the narrative or initial-world facts. Mechanical checks cover document structure, identifiers, references, causal-graph consistency, and tier-specific construction boundaries. Evaluation results do not rewrite a frozen case version.

\paragraph{Deduction cases}

Authors specify the map, entities, initial conditions, timed processes, action rules, ordered outcomes, and background maintenance obligations. They also provide a zero-intervention timeline as an implementation-validation target. Case modules transcribe these dynamics through the generic engine's case hooks. Zero-intervention replay checks agreement with the authored timeline, and an engine-informed oracle checks prediction-scoring consistency. Rule audits, always-wait and sampled random-action baselines, witness trajectories, and development calibration provide additional diagnostics. The frozen dynamics define answers for interventions, which are computed when the corresponding trajectories are evaluated.

\section{Runtime and Reproducibility Details}
\label{app:runtime}

\subsection{Maintenance Representation and Execution}
\label{app:maintenance-runtime}
The maintenance state is $s=(G,Y,I,L)$: $G$ contains regions, tiles, and their connections; $Y$ defines entity record types; $I$ contains typed entities; and $L$ contains reusable capabilities. An entity's holder may be a tile, another entity, or null, with non-cyclic holder chains resolving spatial containment when a tile is reached. Each capability describes a behavior and performs at most one field-level write, or no write for a recorded behavior without a persistent effect. This representation is specific to maintenance. Rendering visualizes the resulting world but is not itself a scored output.

A runner exposes the world interface over the Model Context Protocol~\citep{mcp2024}, delivers successive narrative inputs, and controls round boundaries. The agent can retrieve all narrative inputs released so far and query current authoritative data, including world logs from the start of the run to the current operation. Query access only returns the corresponding fragment of the query. The returned observations enter its tool-interaction history $h_{t,j}$. Read-only calls are recorded in the interaction audit, while the world log records engine operations and their accepted or rejected outcomes.

\begin{table}[h]
  \caption{The six world-interface tools exposed to the maintenance agent. Construction permissions depend on difficulty tier (Section~\ref{sec:maintenance}).}
  \label{tab:tools}
  \centering
  \begin{tabular}{ll}
    \toprule
    Tool & Function \\
    \midrule
    \texttt{help} & return the public protocol, data schemas, and construction rules \\
    \texttt{get\_context} & return the cumulative narrative, run coordinates, and remaining budget \\
    \texttt{read\_data} & read-only paginated queries over the state and the log \\
    \texttt{construct} & create, update, or delete one type, entity, or capability \\
    \texttt{commit\_step} & submit one or more capability triggers as one atomic step \\
    \texttt{end\_round} & close the current round and freeze its snapshot \\
    \bottomrule
  \end{tabular}
\end{table}

The engine enforces transaction-level atomicity for each trigger batch: the triggers are evaluated against the same pre-state and accepted or rejected as a whole. It checks format, references, holder constraints, write conflicts, and the resulting structural validity in a staged world copy. If any check fails, the entire batch is rejected without applying any proposed world changes; otherwise, all triggers are recorded and their optional writes are committed together. This all-or-nothing execution rule does not establish that a batch represents one semantically atomic behavior or a physically plausible process, nor does it guarantee crash-atomic persistence across world files, logs, and run metadata. Invalid requests receive a generic rejection rather than a detailed diagnosis, and rejection does not directly zero the run's score.

Organizing coherent event flows remains the agent's responsibility. It must explicitly trigger character actions, NPC behaviors, and environmental changes; the engine does not autonomously schedule background events, and narrative correctness is evaluated afterwards. Within an accepted batch, triggers represent concurrent events, and their submission order carries no meaning. Dependent stages must therefore be realized through separate triggers in appropriately ordered steps. Step is a discrete index of accepted batches, not a physical clock: each accepted batch advances it by one, whereas construction operations and rejected batches do not. Engine-operation sequence numbers separately order all world operations. When a round's interaction budget is exhausted, the round is closed and scored as it stands.

Before a round opens, the runner requests any author-specified graph additions and permitted presets through the engine. Hard permits agent construction of capabilities and does not inject new capabilities in later rounds; Extreme additionally delegates types and entities and does not inject those records in later rounds. Pre-existing foundations belong to the initial data. Round initialization is part of $F_A$ in Eq.~\eqref{eq:maintenance}; $h_{t,j}$ contains only what the agent has actually received or queried. Closing a round, whether explicitly or on budget exhaustion, freezes its state and log boundary before later rounds proceed.

A read-only replayer reapplies logged changes to frozen initial data and checks reconstructed world states and Step values against frozen round snapshots. The benchmark agent runs outside the authoritative world store, with world access through the runner and a separate workspace for its own files and memory. Runtime configuration and execution artifacts are saved for replay. For both tracks, replaying saved records does not imply that fresh model calls will produce identical trajectories.

\subsection{Deduction Dynamics and Interaction}
\label{app:deduction-runtime}

Deduction uses a separate deterministic engine with case-defined actors, spatial fields, activities, journeys, and an internal clock. Case hooks determine action effects, NPC choices, timed processes, milestones, cascades, and outcomes. These define the dynamics $f$; the broader execution mechanism $F$ also specifies the interaction and information-access rules below. The tested model neither constructs maintenance capabilities nor edits world records. M2 gives it control of one actor, whereas M1 supplies no intervention. Thus the common notation in Section~\ref{sec:task} does not imply a shared concrete engine or state schema.

An M2 action advances the engine to the next decision point: action completion, interruption, or a per-tile checkpoint when the agent selects stepwise movement. Other actors and background processes continue during this interval, unless a terminal outcome ends execution earlier. Rejected actions also advance the clock, for up to the case's configured wait duration or until an earlier terminal outcome. Feedback distinguishes syntax errors from failures to ground a valid command in the current world. A prediction is recorded before its corresponding result is revealed and never changes the reference trajectory.

M1 submission ends when the model declares an outcome in its prediction or reaches the public segment limit (the case horizon divided by the segment length, rounded up). These stopping conditions do not disclose the true terminal time. M2 ends at a world outcome or the configured maximum number of decisions.

Both information conditions provide the initial public fields. M1-Open gives detailed process rules, whereas M1-Naming provides qualitative process descriptions without all numeric parameters. M2 shares a process description across conditions: Open additionally exposes machine-readable action rules, while Naming exposes an action vocabulary with brief descriptions and allows inference about its use from state feedback and rejection reasons. M1 has no true intermediate-state feedback, so unspecified parameters cannot generally be identified from observation: its Naming results partly measure prior knowledge and guessing, not feedback-based rule induction.

Deterministic deduction replay requires the same engine and case code versions, the same initial conditions, and the recorded action sequence. Current runtime metadata records the case identifier, model settings, harness version, and image name, but does not automatically fingerprint the engine or case source. The corresponding source revision must therefore be preserved separately; a saved run directory alone is not a self-contained versioned copy of the dynamics.

\section{Contract Items, Judging Worksheet, and Audit Invariants}
\label{app:scoring}

\begin{table}[h]
  \caption{The six contract item types of the maintenance track. Semantic items are judged item by item by an isolated language-model judge. Authored event relations and both scores are computed by programs from validated semantic bindings. A structure requirement enters the completion score only in tiers that assign its construction to the agent, and type-level requirements are audited but never scored.}
  \label{tab:items}
  \centering
  \begin{tabular}{@{}clp{6.5cm}l@{}}
    \toprule
    \multicolumn{2}{l}{Item} & What it checks & Settled by \\
    \midrule
    $E$ & event & a contracted event happened, bound to the accepted triggers that realized it & judge + audit \\
    $A$ & structure & a required structure was prepared and satisfied its availability and use constraints & judge + audit \\
    $P$ & probe & nothing forbidden occurred while the precondition held & judge + audit \\
    $CA$ & causal & all authored causes and the effect occurred, with each cause ending before the effect begins & program \\
    $TM$ & temporal & one event must precede another, immediately or not & program \\
    $CC$ & concurrency & two authored event intervals start together, overlap, or end together & program \\
    \bottomrule
  \end{tabular}
\end{table}

Maintenance scoring has three stages. An isolated LLM session for each round fills a prefilled worksheet containing the contract items and current-round candidate triggers. Deterministic checks then validate field shapes, round identity, reference ranges, trigger attribution, and structure lifecycles. Finally, programs derive event intervals, merge cross-round A/P fragments, settle authored relations, and compute scores. A failed semantic item receives a failed result; a structurally invalid scoring artifact instead rejects the evaluation attempt and produces no new score. Presentation-only failures do not invalidate an otherwise valid score.

The judge may read the complete frozen GT document set as its criterion, but its world evidence is the current round's frozen snapshot and the operation-log prefix ending at that round's boundary. Each session uses an isolated judge profile and may edit only its designated worksheet. The read boundary is specified by the manifest and tool-use protocol; it is not claimed to be a filesystem sandbox against a malicious judge. A help tool exposes the worksheet schema, and a verify tool runs the same local deterministic checks used by the evaluator.

\paragraph{Event realization and attribution.}
An E item must bind one or more accepted triggers from its own round. For a persistent effect, the judge checks the actual targets, inputs, and before/after values against the required behavior and result; the capability's name alone is insufficient. For a no-write behavior, the triggered capability's description and inputs supply the behavioral evidence. A multi-step event must bind all triggers jointly realizing it rather than only the final write. Independent prerequisites belong to their own events and cannot be reused as evidence for another E. A single macro-trigger that bundles multiple independent required events fails all affected E items, even when their final states appear correct. A multi-trigger atomic batch is not itself such a macro: its distinct triggers can realize distinct concurrent events.

\paragraph{Structure availability and use.}
Logic-A checks semantic capability preparation and actual use on each declared A-to-E edge. One authored requirement may map to several finer capabilities, but a generic writer cannot stand in for multiple semantically distinct Logic-A requirements. Item-A requires one-to-one object identity; Type-A permits broader or finer type mappings and is diagnostic only. A qualifying structure prepared in an earlier round can satisfy a later requirement. For an established E, each used Logic generation must have been prepared before that E's earliest bound trigger and remain available at its own trigger positions. For an unestablished E, a matching prepared capability must survive at the round boundary. Item/Type generations must span their dependent events' bound intervals, using the round boundary when an event is unestablished; Item-A additionally preserves the same identity and generation across fragments. Whole-record updates through construction, and deletion/recreation, begin new generations; ordinary field writes by triggers do not. A failing A does not change the E result, and a missing E does not automatically invalidate a correctly prepared A.

\paragraph{Probe scope.}
Each P specifies an activation condition and forbidden behavior over a declared scope. The judge inspects the fragment's entry state and effective content operations, testing the condition at the corresponding time. Eligible operations are accepted round initialization, structure construction/update/deletion, and trigger batches, including no-write triggers. It returns entry violations and identifiers of violating operations, with concise reasons. Programs validate that reported operations belong to the scope and are eligible; queries, rejected requests, and round/run-control operations are not violation evidence. An empty violation list means that the judge found none, not that software has proved exhaustive semantic review. Prohibitions on routes, means, or conditions are assessed independently from whether an event reached its requested result.

\paragraph{Intervals and authored relations.}
For an established event $e$, let $S_e$ be the nonempty set of successful world Steps containing its bound triggers. Its interval is $[\alpha_e,\beta_e]=[\min S_e,\max S_e]$, where $\alpha_e$ and $\beta_e$ are its start and end. Every relation requires all referenced events to be established. For an effect $v$ with authored direct-cause set $\operatorname{pa}(v)$, CA requires $\max_{u\in\operatorname{pa}(v)}\beta_u<\alpha_v$. TM from $u$ to $v$ requires either $\beta_u<\alpha_v$ or $\alpha_v=\beta_u+1$ for immediate succession. A CC interval is bounded by the start of its declared starting event and the end of its declared ending event. For two valid intervals $[a,b]$ and $[c,d]$, CC checks $a=c$ (same start), $b=d$ (same end), or $\max(a,c)\leq\min(b,d)$ (closed-interval overlap). Adjacent nodes $u,v$ in an event chain require $\beta_u<\alpha_v$. CA dependencies are authored semantic requirements; their programmatic check establishes event presence and ordering, not causality inferred from temporal correlation or interventions.

\paragraph{Cross-round settlement and denominators.}
Events remain assigned to their own rounds. A and P fragments are merged by global ID; any failing fragment makes the corresponding global item fail, and that item enters completion only once, in its final fragment round. Only agent-assigned Logic-A and Item-A enter $A^{\mathrm{sc}}_{t,d}$; pre-built structures and all Type-A remain audited but unscored. Relations enter the round of their consequent or contract declaration. The implementation retains both passed and total counts by item family, returns 1 for a score with an empty denominator, and defines no mixed completion/relation score or default cross-round weighted total. Aggregate comparisons therefore need an explicitly stated macro/micro convention and must distinguish the effect of changing A denominators across difficulty tiers.

\section{Deduction Targets, Readouts, and Reference Baselines}
\label{app:deduction-scoring}

To obtain the answer $y_j$ defined in Eq.~\eqref{eq:deduction}, the benchmark uses the query's true starting state $s_j$, specified action sequence $u_j$, and query specification $q_j$:
\begin{equation}
y_j=\operatorname{Ref}_f(s_j,u_j;q_j).
\label{eq:deduction-reference}
\end{equation}
The reference operator executes the fixed dynamics $f$ and extracts the requested answer. This computation uses the true state without disclosing it to the tested model. Targets use public scoreable fields rather than internal engine counters. For M1 segments and M2 interval/lookahead queries, each field target contains the path and end value of a field that differs from the starting state. Counterfactual targets instead compare alternative and actual trajectories at the same query time. Rule-firing predictions name the rule IDs that occurred, while outcome predictions identify the ending and, for M1, its time. These answers summarize reference execution; they need not reproduce every internal record.

\paragraph{M1 autonomous rollout.}
The engine generates a zero-intervention trajectory and divides it into segments of $k$ world-time units, truncating the final segment at termination. In Eq.~\eqref{eq:deduction-reference}, $s_j$ is the true state at the start of the queried segment, $u_j$ specifies autonomous waiting, and $q_j$ requests that segment's changes and event/outcome information. The model predicts from the initial public state, disclosed dynamics, and its earlier predictions; the engine's later $s_j$ states and correctness feedback are not returned. Missing trailing predictions are scored as empty predictions. The per-segment field-F1 curve shows how errors vary with rollout depth, and episode field F1 pools matched, predicted, and actual field-change counts.

\paragraph{M2 action-interval prediction.}
Here $s_j$ is the current pre-action state, $u_j$ contains the submitted action, and $q_j$ ends at the next decision point. The action and prediction are submitted together, before execution feedback is revealed. The target covers all changes over that interval, including autonomous processes and any elapsed time after rejection, rather than only direct effects of the chosen action.

\paragraph{M2 lookahead.}
Each decision also supplies an action plan and predicts changes from the current state to $k$ world-time units later. The reference starts at that state's clone, uses the plan as $u_j$, and stops at the horizon specified by $q_j$ or at earlier world termination. Autonomous waiting fills any time remaining after the plan ends. This plan defines a forecast target, not a commitment to execute it in the live world.

\paragraph{M2 counterfactual prediction.}
Questions are generated after milestones and every ten decisions when a legal alternative is found in the bounded candidate set. They replace the most recent chosen action and request the field differences from the actual world at the same query time. For this query, $s_j$ is the state immediately before intervention and $u_j$ contains the alternative action followed by any recorded subsequent commands, without replanning. The query specification $q_j$ identifies the comparison time and actual trajectory used as the reference baseline. Both trajectories are compared at that time, or at their terminal states if they end earlier. Each reported pair gives a differing field's value in the alternative trajectory. These are alternative-versus-actual differences, not the start-versus-end changes requested by interval and lookahead predictions.


\paragraph{Exact field matching.}
For a well-formed prediction, each field appears at most once. For query $j$, let $D_j^{\mathrm{pred}}$ and $D_j^{\mathrm{ref}}$ be the predicted and reference sets of field--value pairs, $H_j=|D_j^{\mathrm{pred}}\cap D_j^{\mathrm{ref}}|$ their exact matches, and $N_j^{\mathrm{pred}}=|D_j^{\mathrm{pred}}|$, $N_j^{\mathrm{ref}}=|D_j^{\mathrm{ref}}|$ their sizes. The pooled field score for one readout is
\begin{equation}
\mathrm{F1}
=\frac{2\sum_j H_j}
{\sum_j N_j^{\mathrm{pred}}+\sum_j N_j^{\mathrm{ref}}}.
\label{eq:deduction-f1}
\end{equation}
Queries with neither predicted nor actual changes have no field-F1 value and do not contribute to the pooled counts. A nonempty reference with an empty prediction scores zero, as does predicting changes when none occurred. If the pooled denominator is zero, the score is undefined rather than a free perfect score. M1 retains vacuous segments in the reported segment count; M2 reports action-interval, lookahead, and counterfactual F1 separately. Rule-firing F1 and outcome matching are secondary diagnostics; no language-model judge is involved.

Rule-firing F1 compares sets of rule IDs and is averaged over intervals with a defined rule score; M1 further excludes field-vacuous segments from this diagnostic. M1 outcome checks use the first declared outcome, comparing its identity and reported time with the reference ending. M2 reports the fraction of action intervals with a matching outcome value, including intervals where both values are null; this diagnostic is distinct from successfully achieving a goal.

M2 also reports task outcome, elapsed world time, accepted/rejected action counts (B1), and case-defined background maintenance with omission costs (B2). These are separate readouts, not a composite score, and the implementation reports raw outcomes rather than automatically subtracting a zero-intervention score. The current defaults are $k=48$ world-time units for M1 segments and M2 lookahead, with a tolerance of 12 units for M1 outcome timing. Their sensitivity requires empirical testing; it is not established by the deterministic engine.

\paragraph{Baselines and their scope.}
The static predictor declares no field changes: its field F1 is zero whenever any reference changes are scored, and undefined if the entire reference is empty. This baseline concerns field prediction, not every secondary readout. An oracle predictor obtains answers from the engine and should score one on non-vacuous targets, checking the prediction-scoring pipeline. Always-wait, sampled random legal actions, and authored witness trajectories provide distinct references for action selection and outcomes. A witness is best-known or feasible unless optimality has separately been proved. Reproducing an authored zero-intervention timeline tests that particular trajectory, while oracle agreement tests scoring against the implementation; neither alone proves that all branches implement every intended world rule.

\section{Full-Benchmark Results for the Three Complete Models}
\label{app:full-benchmark}

DeepSeek v4.1 Flash, Qwen-3.8 Flash, and GLM 5.3 Flash ran every case of both tracks, with 93 maintenance runs each (31 cases $\times$ three tiers) and 120 deduction episodes each. The main text confines cross-model maintenance claims to the nine-case common grid of Appendix~\ref{app:subset-selection}, on which every maintenance model ran identical cases, while this appendix reports the full-grid detail. All numbers are recomputed from the saved per-run records. Paired comparisons use two-sided Wilcoxon signed-rank tests, pairing the 93 runs of two models on maintenance (or the 31 cases within a tier) and the family-level cells defined below on deduction.

\subsection{Maintenance Track on the Full 31-Case Grid}
\label{app:full-maintenance}

\begin{table}[p]
  \caption{Per-case completion (C) and relation (R) scores on the full maintenance grid, 93 runs per model. DS = DeepSeek v4.1 Flash, QF = Qwen-3.8 Flash, GF = GLM 5.3 Flash. $^{\dagger}$ marks the single full-crash run discussed in the text.}
  \label{tab:full-maintenance}
  \centering
  \scriptsize
  \setlength{\tabcolsep}{2pt}
  \begin{tabular}{@{}l *{18}{c}@{}}
    \toprule
    & \multicolumn{6}{c}{Easy} & \multicolumn{6}{c}{Hard} & \multicolumn{6}{c}{Extreme} \\
    \cmidrule(lr){2-7} \cmidrule(lr){8-13} \cmidrule(lr){14-19}
    Case & \multicolumn{2}{c}{DS} & \multicolumn{2}{c}{QF} & \multicolumn{2}{c}{GF} &
           \multicolumn{2}{c}{DS} & \multicolumn{2}{c}{QF} & \multicolumn{2}{c}{GF} &
           \multicolumn{2}{c}{DS} & \multicolumn{2}{c}{QF} & \multicolumn{2}{c}{GF} \\
    \cmidrule(lr){2-3} \cmidrule(lr){4-5} \cmidrule(lr){6-7}
    \cmidrule(lr){8-9} \cmidrule(lr){10-11} \cmidrule(lr){12-13}
    \cmidrule(lr){14-15} \cmidrule(lr){16-17} \cmidrule(lr){18-19}
    & C & R & C & R & C & R & C & R & C & R & C & R & C & R & C & R & C & R \\
    \midrule
    Ark\_001 & 0.99 & 0.89 & 0.99 & 0.68 & 0.85 & 0.44 & 0.67 & 0.30 & 0.74 & 0.75 & 0.04$^{\dagger}$ & 0.00 & 0.49 & 0.13 & 0.62 & 0.36 & 0.51 & 0.16 \\
    Ark\_002 & 0.93 & 0.93 & 0.75 & 0.57 & 0.90 & 0.88 & 0.82 & 0.75 & 0.85 & 0.86 & 0.85 & 0.75 & 0.83 & 0.82 & 0.94 & 0.92 & 0.81 & 0.75 \\
    Ark\_003 & 0.91 & 0.85 & 0.94 & 0.96 & 0.95 & 0.96 & 0.71 & 0.68 & 0.87 & 0.94 & 0.90 & 0.86 & 0.74 & 0.74 & 0.79 & 0.87 & 0.83 & 0.72 \\
    Ark\_004 & 0.85 & 0.70 & 1.00 & 1.00 & 0.93 & 0.92 & 0.74 & 0.56 & 0.51 & 0.30 & 0.48 & 0.24 & 0.84 & 0.74 & 0.87 & 0.83 & 0.88 & 0.84 \\
    Ark\_005 & 0.94 & 0.92 & 0.98 & 0.96 & 0.86 & 0.76 & 0.80 & 0.74 & 0.94 & 0.95 & 0.93 & 0.87 & 0.80 & 0.74 & 0.83 & 0.82 & 0.74 & 0.75 \\
    Ark\_006 & 0.83 & 0.61 & 0.99 & 1.00 & 0.85 & 0.67 & 0.58 & 0.43 & 0.88 & 0.80 & 0.78 & 0.54 & 0.68 & 0.51 & 0.84 & 0.86 & 0.83 & 0.78 \\
    Ark\_007 & 0.88 & 0.76 & 0.90 & 0.79 & 0.92 & 0.85 & 0.71 & 0.63 & 0.75 & 0.79 & 0.82 & 0.75 & 0.72 & 0.42 & 0.91 & 0.88 & 0.80 & 0.60 \\
    Chen\_003 & 1.00 & 0.98 & 1.00 & 1.00 & 1.00 & 1.00 & 0.85 & 0.77 & 0.73 & 1.00 & 1.00 & 0.96 & 0.90 & 0.94 & 0.73 & 0.92 & 0.85 & 0.94 \\
    Li\_001 & 0.91 & 0.83 & 0.89 & 0.84 & 0.88 & 0.81 & 0.55 & 0.55 & 0.63 & 0.81 & 0.84 & 0.69 & 0.75 & 0.56 & 0.79 & 0.76 & 0.71 & 0.59 \\
    Li\_002 & 0.94 & 0.89 & 0.94 & 0.89 & 0.94 & 0.86 & 0.53 & 0.62 & 0.86 & 0.77 & 0.69 & 0.53 & 0.54 & 0.51 & 0.60 & 0.59 & 0.55 & 0.35 \\
    Li\_003 & 0.93 & 0.83 & 0.86 & 0.72 & 0.94 & 0.82 & 0.64 & 0.39 & 0.17 & 0.16 & 0.77 & 0.51 & 0.62 & 0.41 & 0.55 & 0.52 & 0.59 & 0.40 \\
    Li\_004 & 0.89 & 0.76 & 0.89 & 0.77 & 0.87 & 0.73 & 0.58 & 0.44 & 0.66 & 0.66 & 0.56 & 0.45 & 0.63 & 0.47 & 0.61 & 0.43 & 0.62 & 0.39 \\
    Li\_005 & 0.86 & 0.64 & 0.33 & 0.26 & 0.85 & 0.59 & 0.84 & 0.55 & 0.80 & 0.62 & 0.84 & 0.64 & 0.56 & 0.28 & 0.69 & 0.50 & 0.60 & 0.36 \\
    Li\_006 & 0.89 & 0.70 & 0.82 & 0.54 & 0.86 & 0.53 & 0.62 & 0.33 & 0.66 & 0.46 & 0.71 & 0.36 & 0.53 & 0.16 & 0.46 & 0.11 & 0.41 & 0.10 \\
    Li\_007 & 0.72 & 0.43 & 0.45 & 0.26 & 0.77 & 0.55 & 0.47 & 0.21 & 0.60 & 0.41 & 0.33 & 0.18 & 0.48 & 0.22 & 0.50 & 0.26 & 0.39 & 0.08 \\
    Li\_008 & 0.86 & 0.68 & 0.91 & 0.79 & 0.92 & 0.80 & 0.84 & 0.46 & 0.80 & 0.55 & 0.93 & 0.76 & 0.79 & 0.58 & 0.70 & 0.43 & 0.43 & 0.22 \\
    Li\_009 & 0.96 & 0.91 & 0.41 & 0.29 & 0.96 & 0.88 & 0.82 & 0.59 & 0.81 & 0.70 & 0.94 & 0.89 & 0.62 & 0.32 & 0.81 & 0.66 & 0.82 & 0.68 \\
    Li\_010 & 0.31 & 0.23 & 0.30 & 0.21 & 0.40 & 0.25 & 0.33 & 0.20 & 0.32 & 0.18 & 0.38 & 0.27 & 0.32 & 0.13 & 0.47 & 0.26 & 0.38 & 0.17 \\
    Li\_011 & 0.68 & 0.43 & 0.75 & 0.56 & 0.85 & 0.60 & 0.51 & 0.25 & 0.16 & 0.12 & 0.16 & 0.12 & 0.37 & 0.11 & 0.42 & 0.18 & 0.37 & 0.13 \\
    Li\_012 & 0.89 & 0.78 & 0.94 & 0.85 & 0.88 & 0.69 & 0.84 & 0.72 & 0.46 & 0.30 & 0.83 & 0.60 & 0.62 & 0.40 & 0.59 & 0.34 & 0.70 & 0.47 \\
    Li\_013 & 0.95 & 0.86 & 0.98 & 0.93 & 0.97 & 0.97 & 0.85 & 0.76 & 0.91 & 0.89 & 0.88 & 0.77 & 0.73 & 0.57 & 0.82 & 0.81 & 0.61 & 0.35 \\
    Li\_014 & 0.93 & 0.84 & 0.94 & 0.87 & 0.90 & 0.80 & 0.60 & 0.51 & 0.27 & 0.21 & 0.59 & 0.46 & 0.67 & 0.50 & 0.65 & 0.51 & 0.57 & 0.27 \\
    Li\_015 & 0.93 & 0.83 & 0.93 & 0.80 & 0.94 & 0.86 & 0.84 & 0.85 & 0.83 & 0.81 & 0.85 & 0.79 & 0.76 & 0.65 & 0.87 & 0.76 & 0.87 & 0.75 \\
    Li\_016 & 0.97 & 0.90 & 0.98 & 0.95 & 0.98 & 0.91 & 0.64 & 0.74 & 0.65 & 0.91 & 0.74 & 0.70 & 0.60 & 0.50 & 0.61 & 0.53 & 0.57 & 0.34 \\
    Li\_017 & 0.97 & 0.91 & 0.97 & 0.92 & 0.95 & 0.90 & 0.64 & 0.74 & 0.76 & 0.80 & 0.84 & 0.76 & 0.69 & 0.53 & 0.75 & 0.70 & 0.76 & 0.64 \\
    Li\_018 & 0.88 & 0.76 & 0.93 & 0.84 & 0.93 & 0.81 & 0.63 & 0.56 & 0.66 & 0.64 & 0.75 & 0.63 & 0.70 & 0.45 & 0.77 & 0.58 & 0.81 & 0.58 \\
    Myth\_001 & 0.80 & 0.66 & 0.83 & 0.74 & 0.91 & 0.85 & 0.81 & 0.77 & 0.81 & 0.85 & 0.80 & 0.77 & 0.72 & 0.55 & 0.82 & 0.74 & 0.76 & 0.65 \\
    Myth\_002 & 0.79 & 0.58 & 0.79 & 0.59 & 0.77 & 0.57 & 0.56 & 0.47 & 0.71 & 0.64 & 0.80 & 0.75 & 0.66 & 0.46 & 0.76 & 0.67 & 0.69 & 0.45 \\
    Myth\_003 & 0.94 & 0.92 & 0.94 & 0.93 & 0.96 & 0.92 & 0.83 & 0.81 & 0.90 & 0.84 & 0.81 & 0.85 & 0.75 & 0.71 & 0.82 & 0.86 & 0.82 & 0.75 \\
    Myth\_004 & 0.94 & 0.78 & 0.45 & 0.18 & 0.69 & 0.43 & 0.73 & 0.61 & 0.78 & 0.55 & 0.61 & 0.29 & 0.71 & 0.44 & 0.94 & 0.85 & 0.80 & 0.53 \\
    Myth\_005 & 0.85 & 0.64 & 0.84 & 0.67 & 0.85 & 0.65 & 0.80 & 0.59 & 0.71 & 0.54 & 0.54 & 0.27 & 0.70 & 0.48 & 0.80 & 0.71 & 0.61 & 0.38 \\
    \midrule
    Tier mean & 0.875 & 0.756 & 0.826 & 0.720 & 0.879 & 0.750 & 0.691 & 0.567 & 0.684 & 0.639 & 0.710 & 0.581 & 0.663 & 0.485 & 0.720 & 0.620 & 0.668 & 0.489 \\
    \bottomrule
  \end{tabular}
\end{table}

On the full grid, completion again fails to separate the three models, clustering within $0.01$ of their nine-case means in Table~\ref{tab:main-results}. Relation remains the discriminating score. Qwen-3.8 Flash leads significantly on relation, DeepSeek v4.1 Flash and GLM 5.3 Flash do not separate on either score, and the only other significant contrast is a completion shift small enough to be practically negligible. All pairwise statistics are reported in Table~\ref{tab:full-maintenance}. The relation advantage of Qwen-3.8 Flash is tier-structured. It is absent in the easy tier, appears on both scores in the extreme tier, and appears on relation in the hard tier.

The difficulty structure of Section~\ref{sec:exp-maintenance} reproduces on the full grid. Completion falls significantly from easy to hard for all three models, and relation falls for DeepSeek v4.1 Flash and GLM 5.3 Flash but only marginally for Qwen-3.8 Flash. From hard to extreme, completion is flat, while relation keeps decaying for the same two models and holds for Qwen-3.8 Flash. Within every tier, both scores decay over rounds, and every round-one-to-round-five decline is significant over the 66 runs with at least five rounds. The round-six uptick reflects case composition, since only Myth-family cases run six rounds. Finally, case-level means correlate strongly across models (Pearson $r = 0.72$--$0.88$ over the 31 cases), showing that which cases are hard is a property of the benchmark rather than of a single model.

Component readouts match Section~\ref{sec:exp-maintenance} as well. Causal checking (CA) is the weakest component for every model in every tier. GLM 5.3 Flash leads hard-tier construction by a wide margin but posts the lowest event score in the same tier and the lowest temporal and concurrency scores at the extreme tier, so its construction strength comes at a visible execution and ordering cost. Qwen-3.8 Flash owns the highest extreme-tier event score.

Per-round derailment scanning (Section~\ref{sec:exp}) separates the models more than any aggregate score. DeepSeek v4.1 Flash completes all 93 runs without derailment. Qwen-3.8 Flash derails in seven runs, all in the Li family, and additionally crashes once at round one with full recovery. Its easy \texttt{Li\_003} run scores zero in the first round and at least $0.90$ in every later round while still averaging $0.86$ overall, so the aggregate alone hides the crash. GLM 5.3 Flash derails twice and crashes once without recovery, namely the \texttt{Ark\_001} hard run marked in Table~\ref{tab:full-maintenance}, whose scores are the lowest of the 279 full-grid runs. The stress case further separates failure forms. DeepSeek v4.1 Flash reads the easy tier best but decays steeply as structure is removed, Qwen-3.8 Flash retains the most relation structure at the extreme tier, and GLM 5.3 Flash collapses at hard and only partially recovers at extreme.

\subsection{Deduction Track: Per-Case Breakdown}
\label{app:full-deduction}

\begin{table}[t]
  \caption{Field-level F1 per case family on the deduction track, reported as open/naming, five-fold episode means. M1 rows report pure-deduction segment F1; M2 rows report closed-loop action-interval F1. Each model ran its own scenario variant of every family (DeepSeek v4.1 Flash the alternate variants such as Ashspring-chip, the other two models the base variants); variants are merged within a family.}
  \label{tab:full-deduction}
  \centering
  \footnotesize
  \begin{tabular}{@{}llccc@{}}
    \toprule
    Case family & Mode & DeepSeek v4.1 Flash & Qwen-3.8 Flash & GLM 5.3 Flash \\
    \midrule
    Ashspring & M1 & 1.00/0.20 & 0.67/0.04 & 0.88/0.16 \\
    Ashspring & M2 & 0.67/0.57 & 0.44/0.41 & 0.75/0.48 \\
    \addlinespace
    Orbit & M1 & 0.96/0.08 & 0.62/0.09 & 0.92/0.07 \\
    Orbit & M2 & 0.80/0.57 & 0.79/0.47 & 0.78/0.68 \\
    \addlinespace
    Manor & M1 & 0.66/0.03 & 0.46/0.05 & 0.79/0.04 \\
    Manor & M2 & 0.30/0.26 & 0.28/0.14 & 0.27/0.17 \\
    \addlinespace
    Maplebrook & M1 & 0.56/0.12 & 0.53/0.08 & 0.61/0.13 \\
    Maplebrook & M2 & 0.41/0.35 & 0.38/0.32 & 0.38/0.36 \\
    \addlinespace
    Siege & M1 & 0.55/0.21 & 0.42/0.11 & 0.30/0.03 \\
    Siege & M2 & 0.74/0.71 & 0.71/0.38 & 0.78/0.60 \\
    \addlinespace
    Spy & M1 & 0.34/0.01 & 0.67/0.03 & 0.76/0.00 \\
    Spy & M2 & 0.46/0.41 & 0.40/0.26 & 0.41/0.27 \\
    \bottomrule
  \end{tabular}
\end{table}

Each model ran $6$ case families $\times$ two rollout modes $\times$ two information conditions $\times$ five folds $= 120$ episodes. Table~\ref{tab:full-deduction} adds the per-family view to the model-level means of Table~\ref{tab:main-results}, from which three observations emerge. First, closed-loop difficulty is a benchmark property, whereas pure-deduction difficulty is not. Family-level M2-open interval F1 correlates strongly across models (Pearson $r = 0.84$--$0.98$), but M1-open correlates weakly ($r = 0.28$--$0.71$), and the \texttt{Spy} family flips outright, ranking last for DeepSeek v4.1 Flash but first for GLM 5.3 Flash under M1-open. Second, the naming collapse is a value-level collapse. Under naming, rule-firing F1 remains high while field F1 falls to $0.11$ or below, so the models keep knowing which rules will fire but not the values those rules produce. Third, closed-loop feedback compensates unevenly, recovering about $85\%$ of the open--naming gap for DeepSeek v4.1 Flash, $79\%$ for GLM 5.3 Flash, and only $66\%$ for Qwen-3.8 Flash, which is where its significantly lower M2 interval score comes from. All per-family statistics are reported in Table~\ref{tab:full-deduction}.

Counterfactual F1 dissociates from interval competence at the family level as well. GLM 5.3 Flash is lowest overall and significantly below both others, with near-zero cells such as \texttt{Ashspring} under open, while DeepSeek v4.1 Flash and Qwen-3.8 Flash do not separate.

Behavioral readouts separate the models where F1 does not. All three produce episodes of at least 100 decision steps under naming, with Qwen-3.8 Flash circling to the horizon twice on \texttt{Manor}, GLM 5.3 Flash three times across \texttt{Manor} and \texttt{Ashspring}, and DeepSeek v4.1 Flash once on \texttt{Siege}. Syntax rejections run at similar rates for all three, while grounding rejections vary more widely, from about $1.7$ per episode for DeepSeek v4.1 Flash to about $3.1$ for GLM 5.3 Flash. Case-defined background maintenance is kept in roughly a third of the windows for every model. Recorded token totals reach $182.6$M and $199.4$M for DeepSeek v4.1 Flash and Qwen-3.8 Flash, and M1 episodes cost two to four times less than M2 episodes. The GLM 5.3 Flash deduction batch ran on the dsh harness and carries no token records, so its deduction readings are cross-harness comparisons.

\section{Construction of the Nine-Case Common Subset}
\label{app:subset-selection}

\begin{table}[t]
  \caption{The nine selected cases. Checkpoints count the scored contract items of a case over all three tiers (E, A, P, CA, TM, CC). Spread is the mean over tiers of the cross-model range of the mix score (equal-weight mean of completion and relation) among the three reference models. Consistency counts how many of a case's three tiers reproduce the reference models' difficulty-level ordering (out of 3).}
  \label{tab:subset-selection}
  \centering
  \footnotesize
  \begin{tabular}{@{}llccc@{}}
    \toprule
    Stratum & Case & Checkpoints & Spread & Consistency \\
    \midrule
    Small & \texttt{Myth\_004} & 725 & 0.365 & 2/3 \\
    Small & \texttt{Ark\_006} & 583 & 0.288 & 2/3 \\
    Small & \texttt{Ark\_002} & 472 & 0.167 & 3/3 \\
    \addlinespace
    Medium & \texttt{Li\_009} & 1325 & 0.299 & 3/3 \\
    Medium & \texttt{Li\_002} & 1195 & 0.136 & 2/3 \\
    Medium & \texttt{Myth\_002} & 1097 & 0.143 & 2/3 \\
    \addlinespace
    Large & \texttt{Ark\_001} & 1928 & 0.243 & 3/3 \\
    Large & \texttt{Li\_005} & 1738 & 0.228 & 3/3 \\
    Large & \texttt{Li\_003} & 1730 & 0.204 & 2/3 \\
    \bottomrule
  \end{tabular}
\end{table}

The nine-case grid on which all nine maintenance models are compared (Table~\ref{tab:main-results}) is a calibrated subset of the 31-case library, selected before the six non-reference maintenance models ran. A unit is one case at one difficulty tier, so the full grid has $31 \times 3 = 93$ units and the subset $9 \times 3 = 27$, covering 129 of 444 rounds ($29\%$ of the full-grid cost). Selection used only the full-benchmark models themselves as references, with GLM 5.3 Flash then covering 56 of 93 units, and no other model's results entered the selection.

\paragraph{Stratification by case size.}
Case size is the total number of scored checkpoints over a case's three tiers, $\sum_{d \in \{\mathrm{easy},\mathrm{hard},\mathrm{extreme}\}} \bigl(|E|+|A|+|P|+|CA|+|TM|+|CC|\bigr)$, ranging from $330$ to $3969$ (median $1296$). Cases are sorted by size and split into three strata of $10$, $10$, and $11$ cases. Size rather than genre is used as the stratification variable because checkpoint count measures how many things can go wrong and tracks the score level, whereas genre families mix small and large cases, the Ark family spanning $472$--$1928$ checkpoints. On the reference models, stratifying by family quota missed the full-grid mean by $-0.058$ and $-0.065$ mix points, while size stratification missed by $-0.011$ and $-0.020$.

\paragraph{Filters.}
Four rules are applied per case. (1)~\emph{Ground-truth integrity}. The per-tier item totals must agree across reference models, since they are properties of the frozen case rather than of a run, and all 93 units pass. (2)~\emph{Degenerate runs}. A reference model whose event pass rate on a unit falls below $0.15$ while the others exceed $0.50$ is excluded from that unit's statistics, which occurs once (\texttt{Ark\_001} hard under GLM 5.3 Flash, event rate $0.02$ against $0.84$ and $0.93$). (3)~\emph{Order consistency}. At least two of a case's three tiers must reproduce the reference models' difficulty-level ordering, with DeepSeek v4.1 Flash ahead at easy and Qwen-3.8 Flash ahead at hard and extreme, and eight cases fail, several of them high-spread. (4)~\emph{Within-stratum ranking}. Cases are ranked by mean cross-model spread and the top three per stratum are selected.

\paragraph{Result and validation.}
Table~\ref{tab:subset-selection} lists the nine cases, three from Ark, four from Li, and two from Myth. The procedure deliberately concentrates discriminating cases, with spreads ranging from $0.136$ to $0.365$ (mean $0.230$) and 13 of the 27 units having a spread of at least $0.25$. Two exclusions are deliberate. \texttt{Chen\_003}, the smallest case at $330$ checkpoints, is saturated at $0.90$--$1.00$ for every reference at every tier and carries no discrimination. The three largest cases, \texttt{Li\_011} at $3969$, \texttt{Li\_012} at $2607$, and \texttt{Li\_010} at $2339$ checkpoints, fall in the large stratum but fail order consistency or spread, so the subset covers $472$--$1928$ checkpoints and does not represent behavior beyond that scale. Against the full grid, the subset's mix mean is biased by $-0.011$ (DeepSeek v4.1 Flash) and $-0.020$ (Qwen-3.8 Flash), and the reference models' ordering is preserved. Per tier, the easy tier runs $0.051$ below the full grid, because easy cases saturate above $0.9$ with no spread and the filter trades easy-tier representativeness for discrimination, while hard and extreme match within $0.01$. Consistently, the full-grid completion means of the three references (Appendix~\ref{app:full-maintenance}) agree with their nine-case means in Table~\ref{tab:main-results} to within $0.01$.

\section{Per-Model Component Radars}
\label{app:component-radar}

\begin{figure}[t]
  \centering
  \includegraphics[width=0.92\linewidth]{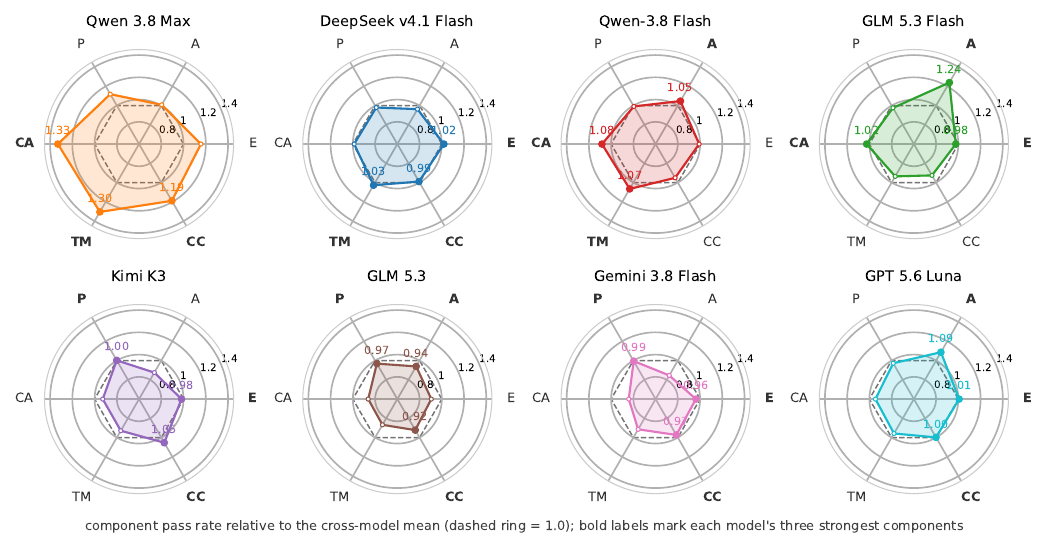}
  \caption{Per-model component radars on the maintenance track (nine-case common grid). Each panel shows one model's per-component pass rate, averaged over the three difficulty tiers and normalized by the cross-model mean; the dashed ring marks the mean (1.0). Bold labels and filled markers denote the model's three strongest components.}
\end{figure}

Figure~\ref{app:component-radar} splits the merged component radar of Figure~\ref{fig:component-radar} into per-model panels. Each panel shows the model's per-component pass rate, averaged over the three difficulty tiers and normalized by the cross-model mean; bold labels and filled markers denote the model's three strongest components. Qwen 3.8 Max is the only model above the mean on all six components; the per-model views make the single-strength profiles of GLM 5.3 Flash (A), Qwen-3.8 Flash (CA, TM), GPT 5.6 Luna (A), and Kimi K3 (CC), and the flat profiles of DeepSeek v4.1 Flash, Gemini 3.8 Flash, and GLM 5.3 directly visible.

\end{document}